**Title:** Enhancing scientific exploration of the deep sea through shared autonomy in remote manipulation

**Authors:**
Amy Phung,[1,2] Gideon Billings,[2] Andrea F. Daniele,[3] Matthew R. Walter,[3] Richard Camilli[2]*

**Affiliations:**
[1]Massachusetts Institute of Technology; Cambridge, MA.
[2]Woods Hole Oceanographic Institution; Falmouth, MA.
[3]Toyota Technological Institute at Chicago; Chicago, IL.

**One-Sentence Abstract:** Shared autonomy enables novice remote users to conduct deep ocean science operations with robotic manipulators.

**Main Text:**

Scientific exploration of the deep ocean is vital for understanding natural Earth processes, but remains inaccessible to most (1). Dexterous sampling operations at depth are typically conducted by robotic manipulator arms onboard remotely operated vehicles (ROVs), which are directly teleoperated by pilots aboard surface support vessels. This presents barriers to access due to the infrastructure, training, and physical ability requirements for at-sea oceanographic research. Enabling shore-based participants to observe and control robotic sampling processes can reduce these barriers; however, the conventional direct-teleoperation approach is infeasible for remote operators due to the considerable bandwidth limitations and latency inherent in satellite communication. Thus, some degree of ROV autonomy is required to support remote operations (2).

To address this need, our team developed the SHared Autonomy for Remote Collaboration (SHARC) framework (3), which enables remote participants to conduct shipboard operations and control robotic manipulators using only a basic internet connection and consumer-grade hardware, regardless of their prior piloting experience. SHARC extends current supervisory control methods by enabling real-time collaboration between multiple remote operators, who can issue goal-directed commands through speech and hand gestures. SHARC couples these natural input modalities with an intuitive three dimensional (3D) workspace representation that segments the workspace and actions into a compact representation of known features, states, and policies. The flexibility of language enables users to succinctly issue complex commands that would otherwise be difficult to execute with conventional controllers. In addition to reducing cognitive load, the intuitive nature of speech and gestures minimizes the training required for operation and makes SHARC accessible to a diverse population of users. These natural input modalities also have the benefit of remaining functional under intermittent, low-bandwidth, and high-latency communications.

The ability to involve remote users during field operations became particularly important during the COVID-19 pandemic, when space onboard research vessels was especially restricted. During an oceanographic expedition in the San Pedro Basin of the Eastern Pacific Ocean, our remote team members operated the Nereid Under Ice (NUI) vehicle (4) from thousands of kilometers away using SHARC's virtual reality (VR) and desktop interfaces. The team collaboratively collected a physical push core sample and recorded in-situ X-ray fluorescence (XRF) measurements of seafloor microbial mats and sediments at water depths exceeding 1000 m, whilst being physically located across the United States in Chicago, Boston, and in the village of Woods Hole, MA (Fig. 1).

XRF sensors have broad utility but are rarely used for in-situ observations underwater because the high attenuation rate of X-rays in water requires the delicate sensor to be in direct contact with the sediment. Prior experimental methods for underwater in-situ XRF measurements limit precision over sample site selection to tens of meters to kilometers (5,6). With SHARC, the remote science team ensured the XRF maintained contact with the seafloor with centimeter-level positioning during sample acquisition, thereby reducing environmental disturbance to the work site and enabling uninterrupted measurement during the sensor integration time without damaging the sensor. SHARC's real-time data feedback enabled scientists to actively tune the XRF's X-ray source and detector parameters to maximize the signal-to-noise ratio during measurement. The resulting spectra revealed elevated concentrations of iron within the microbial mats, which suggested the presence of chemolithoautotrophs (7). To independently determine the presence of these microbes, the remote science team used SHARC to collect a physical push core sample from the same microbial mat.

Conventional methods for deep ocean sampling require operators to mentally construct a 3D scene from a variety of 2D camera feeds, which is particularly challenging when such feeds are low-resolution or framerate-limited. This cognitive load is exacerbated in applications such as underwater intervention, where inadvertent collisions with the environment and vehicle can be catastrophic. In contrast to conventional interfaces, SHARC enables users to operate with performance benchmarks, including precision, accuracy, task time, and task completion rate comparable to that of trained pilots regardless of their prior experience, even when faced with bandwidth limitations. Our field demonstration highlights SHARC's ability to complete delicate operations in unstructured environments under bandwidth-limited conditions, which may be extensible to other sensitive domains where dexterity is required. Applications include nuclear decommissioning (8), deep space operations (9), and unexploded ordnance/disposed military munition remediation (10)).

SHARC's fundamental strategy for task allocation entails delegating responsibilities between the robot and operator based on their complementary strengths. Human operators are responsible for high-level scene understanding, goal selection, and task-level planning, which are challenging for existing perception and decision-making algorithms in unstructured underwater environments. Meanwhile, SHARC relegates capabilities that can readily be solved using autonomy algorithms to the robot. By automating the inverse kinematics, motion planning, low-level control, and obstacle avoidance processes to the robot, SHARC can improve task efficiency (11). Critically, SHARC renders the robot's intended actions prior to execution in context of its understanding of the surrounding environment as a 3D scene reconstruction along with the location and label of detected tools. This visual representation improves upon conventional interfaces by providing visual confirmation of the planned action prior to execution. With this approach, operators no longer need to simultaneously interpret the robot's many high-frequency sensor streams while solving the low-level manipulator kinematics necessary to move the end-effector. Instead, these tasks are offloaded to the robot, which should reduce operators' cognitive load during operation.

Laboratory-based user studies indicate that SHARC enables experienced pilots as well as novices to complete manipulation tasks in less time and greater precision than it takes the pilots to complete the same tasks using a conventional underwater manipulator controller, particularly in bandwidth-limited conditions. These results suggest that SHARC enables efficient human-collaborative manipulation, which can out-perform conventional teleoperation in complex, unstructured environments. By reducing the required bandwidth needed for operation, SHARC can potentially enable tether-less manipulation operations, which would increase the ROV's capabilities while reducing operational costs. SHARC's ability to relax infrastructure requirements enables remote scientists and other novice shore-side users to participate without requiring additional bandwidth from the ship or specialized hardware. Thus, SHARC provides a promising avenue for democratizing access to deep-ocean science and expanding scientific engagement to a broader audience, including classrooms and the general public.

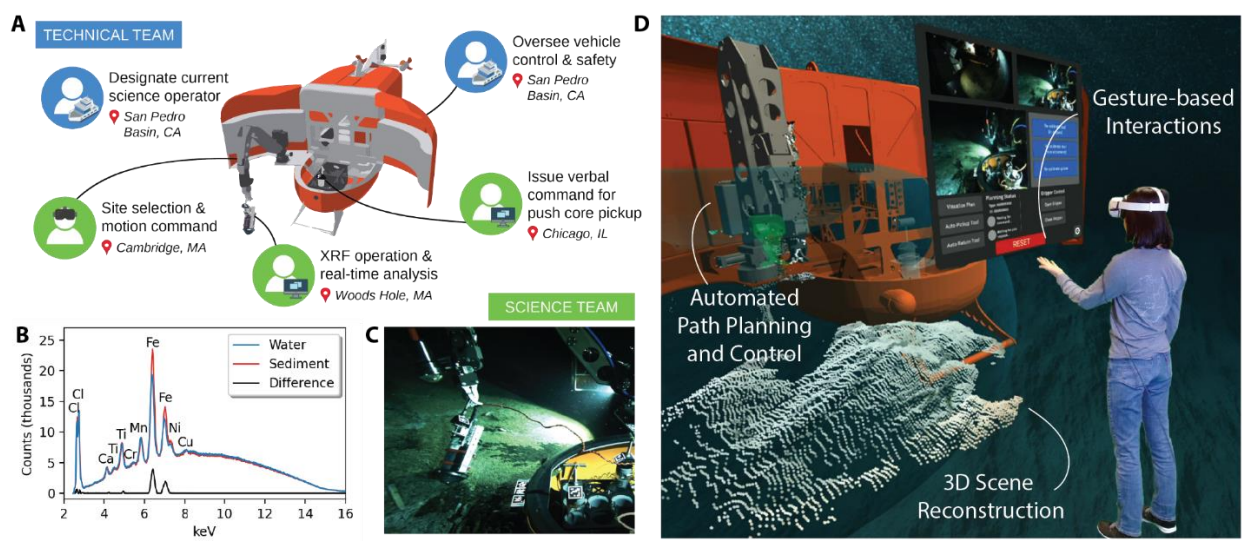

**Fig. 1. Demonstration of shared autonomy with SHARC, conducting deep ocean *in-situ* XRF analysis in real time by a remote science team distributed across North America.** (**A**) Illustration of the sampling process with SHARC. Remote scientists (green) using SHARC-VR (headset icon) and SHARC-desktop (monitor icon) collaborated with the onboard crew (blue) to take an XRF measurement and push core sample of a microbial mat within the San Pedro Basin. (**B**) XRF spectra indicate elevated iron concentrations in the microbial sample (red) above ambient (blue). (**C**) Snapshot of a representative video frame broadcasted with SHARC during measurement. (**D**) From the SHARC-VR interface, users can specify task-level objectives with hand gestures and natural language. SHARC automates the low-level planning and control, enabling safe and robust manipulation over low-bandwidth connections.

**Acknowledgments:** The authors would like to acknowledge primary support from the National Science Foundation National Robotics Initiative which has made this research possible, additional support from NASA's PSTAR program, and in-kind support by the NOAA Ocean Exploration Cooperative Institute with ship and robotic vehicle operations during 2021 Pacific Ocean demonstrations in the San Pedro Basin. The authors would also like to thank the captain and crew of the R/V Nautilus, the NUI robotic vehicle operations team, and study participants who volunteered to assist with performance testing of the SHARC and conventional robotic manipulation systems. AP would like to acknowledge support from the National Science Foundation Graduate Research Fellowship under Grant No. 2141064 and the Link Foundation.

**Funding:**
>National Science Foundation, National Robotics Initiative grant IIS-1830500 (RC)
>National Science Foundation, National Robotics Initiative grant IIS-1830660 (MW)
>National Aeronautics and Space Administration, Planetary Science and Technology from Analog Research grant NNX16AL08G (RC)

**Author contributions:**
   Conceptualization: AFD, AP, GB, MRW, RC
   Methodology: AFD, AP, GB, MRW, RC
   Investigation: AFD, AP, GB, MRW, RC
   Visualization: AFD, AP, GB, RC
   Funding acquisition: MRW, RC
   Project administration: MRW, RC
   Supervision: MRW, RC
   Writing – original draft: AFD, AP, GB, MRW, RC
   Writing – review & editing: AFD, AP, GB, MRW, RC

**Competing interests:** Authors declare that they have no competing interests.

**Data and materials availability:** All data are available in the main text or the supplementary materials.